\documentclass[conference,compsoc]{IEEEtran}

\usepackage[T1]{fontenc}
\usepackage{times}
\usepackage{xcolor}
\usepackage{multirow}
\usepackage{amsmath}
\usepackage{amssymb}
\providecommand{\texttimes}{\ensuremath{\times}}
\providecommand{\degree}{\ensuremath{^\circ}}
\usepackage{booktabs}
\usepackage{threeparttable}

\ifCLASSOPTIONcompsoc
  \usepackage[nocompress]{cite}
\else
  \usepackage{cite}
\fi

\usepackage{graphicx}
\usepackage{epstopdf}

\graphicspath{{./plots/}{./graphics/}}
\DeclareGraphicsExtensions{.eps,.pdf,.png}

\usepackage[unicode=true,
bookmarks=false,
breaklinks=false,
pdfborder={0 0 0},
pdfborderstyle={},
backref=false,
colorlinks=false]
{hyperref}

\newif\ifdraft
\draftfalse
\ifdraft
 \newcommand{\note}[2][red]{\textcolor{#1}{#2}}
 \newcommand{\notedme}[1]{\note[cyan]{[dme: #1]}}
 \newcommand{\notemax}[1]{\note[blue]{[max: #1]}}
 \newenvironment{scaffold}{\color{red}}{}
 
\else
 \newcommand{\note}[2][red]{}
 \newcommand{\notedme}[1]{}
 \newcommand{\notemax}[1]{}

\fi

\makeatletter

\ifCLASSOPTIONcompsoc
  \usepackage[caption=false,font=footnotesize,labelfont=sf,textfont=sf]{subfig}
\else
  \usepackage[caption=false,font=footnotesize]{subfig}
\fi

\makeatother

\begin{document}

\title{Industrial Kinematic Trajectory Model (IKTM):\\Coordinate-Free Autoregressive Generator}

\author{\IEEEauthorblockN{Max Amiri}
\IEEEauthorblockA{School of Computing\\
University of Otago\\
Dunedin, New Zealand\\
mahdi.amiri@postgrad.otago.ac.nz}\and

\IEEEauthorblockN{David Eyers}
\IEEEauthorblockA{School of Computing\\
University of Otago\\
Dunedin, New Zealand\\
david.eyers@otago.ac.nz}

}

\maketitle

\begin{abstract}
Mobility simulation supports logistics, safety, and communications planning in industrial environments such as ports, mines, and airports.
Existing trajectory models, however, rely on absolute coordinates, road-network tokens, or semantic zones: representations that are site-specific and not well suited to unstructured industrial terrain.
We introduce the Industrial Kinematic Trajectory Model (IKTM), a coordinate-free trajectory generator that represents industrial vehicle motion through kinematic sequences (speed and heading change) with no absolute spatial reference. IKTM uses an autoregressive causal transformer with probabilistic mixture heads and extends our prior coordinate-free Markovian model with deep sequence modelling and an explicit duration-conditioning signal.
Trained on one site and evaluated zero-shot on three unseen sites, it matches the small-turn shape of the empirical turn-rate distributions of held-out telematics; across all four sites, the per-site mean Jensen--Shannon divergence over 100 sampling seeds spans ${\approx}$0.035--0.050~bits under an oracle-length duration-target protocol and ${\approx}$0.032--0.042~bits under a fully zero-shot prior-length protocol, with similar ranges whether out-of-distribution (OOD) duration targets are drawn from each held-out site's empirical length distribution or from the Site~A prior.
Both protocols stay above the metric's sampling-noise floor ($\le$0.0049~bits). Paired by site, the oracle-length values are 5.3--6.0$\times$ lower than those of a re-implementation of our prior Markovian model under the same 1~Hz protocol.
Termination is duration-conditioned rather than spatial: rollouts stop on 100\% of trials with a length-tracking error of $+0.0 \pm 0.0$~s against the sampled target (100 of 100 exactly on target; $N{=}100$, $T{=}0.2$, untouched in-distribution test split).
\end{abstract}


\section{Introduction}
\label{sec:introduction}

Industrial environments such as ports, open-pit mines, airports, and construction yards depend on realistic fleet simulation for logistics planning, safety auditing, and communications-network evaluation.
Trajectory generation methods designed for passenger vehicles on public roads assume structured road networks, standard lanes, and traffic signals, whereas industrial vehicles navigate open, largely unstructured areas with no fixed turning radii.

Absolute-coordinate, road-network, and map-based representations are also tied to specific sites: a model trained on one facility's spatial reference cannot transfer to another without re-mapping its environment.
We therefore require \emph{coordinate-free} models that learn kinematic behaviours, specifically how vehicles accelerate, decelerate, and turn relative to their own heading, without reference to geographic space. This representation allows cross-site transfer to be evaluated without remapping spatial coordinates.

Trajectory foundation models have scaled quickly, to billion-trace corpora and to training across tens of cities.
MobilityGPT~\cite{haydari2026mobilitygpt} and MoveGPT~\cite{han2025movegpt} introduce the GPT-style autoregressive approach for mobility.
UniTraj (Zhu et al.)~\cite{zhu2025unitraj} combines billion-scale worldwide Global Positioning System (GPS) traces.
TrajFM~\cite{lin2024trajfm} targets vehicle trajectory transfer, and TrajTok~\cite{xiong2026trajtok} proposes density-adaptive spatial tokenisation for the same class of models.
Yet all of the models surveyed here represent trajectories in absolute or relative Cartesian coordinates, or via road-network tokens and semantic zones, which does not match the domain constraints of industrial settings.
To our knowledge, none of them has been designed for, or evaluated in, off-road industrial environments.

Our previous work introduced a Markovian mobility model for industrial vehicles that predicts both speed and direction changes conditioned on the current speed class, fitted to millions of real telematics events from four industrial sites~\cite{Amiri2024Model}.
While effective as a lightweight simulator, it has three limits: its speed-class state carries only one-step memory, so it cannot capture longer-range kinematic dependencies such as gradual deceleration over tens of seconds; its direction-change statistics are estimated from 10-degree bins of turn magnitude; and it has no trip-termination mechanism.
IKTM builds on this base by replacing single-step speed-class transitions with a causal transformer over a 60-step continuous kinematic history, using Gaussian mixture model (GMM) and von Mises mixture heads for smooth probabilistic generation, and introducing duration conditioning for reliable self-termination.

This work targets three main challenges in coordinate-free autoregressive trajectory generation for industrial vehicles:

\begin{enumerate}
    \item \textbf{Turning-Data Preservation}: Industrial telematics is sparsely and irregularly sampled (median fix interval 5~s; 99th percentile 45~s). Standard linear interpolation between checkpoints yields piecewise-straight paths that concentrate turning at abrupt changes in direction rather than across each segment.
    \item \textbf{Fixed-Point Convergence}: Autoregressive models trained with deterministic regression objectives regress towards the conditional mean at each step. During open-loop rollouts this causes the model to converge to a constant-speed, constant-curvature attractor (a fixed point).
    \item \textbf{Trip Termination}: Without spatial boundaries or explicit length constraints, coordinate-free autoregressive generators receive no direct signal for when a trip should end.
\end{enumerate}

We make four contributions:

\begin{itemize}
    \item A trajectory preprocessing pipeline featuring \textbf{heading-aware curved interpolation} based on cubic B\'{e}zier curves. The curve passes exactly through every retained checkpoint (after drift-cleaning removes GPS flyaways) while restoring continuous turning; the integer-second-resampled curve path reconstructs that checkpoint reference to within sub-millimetre median error.
    \item A \textbf{probabilistic causal decoder-only transformer} that predicts parameters of continuous mixture distributions (a GMM for speed and a von Mises mixture for circular heading changes), supporting varied open-loop trajectories compared with deterministic regression heads.
    \item A \textbf{duration-conditioning} mechanism that combines a normalised remaining-time feature with a learned log-duration prior, enabling reliable, length-controllable self-terminating rollouts.
    \item A \textbf{cross-site zero-shot out-of-distribution (OOD) evaluation} on real telematics from four industrial sites, covering teacher-forced next-step prediction and open-loop turn, speed, duration, and temporal statistics.
\end{itemize}

\section{Background and Related Work}
\label{sec:related_work}

\subsection{Trajectory Foundation Models}

Merten et al.~\cite{merten2025trajfromscratch} demonstrate a minimal end-to-end GPT-2 adaptation for spatiotemporal sequences, providing a practical template for our decoder-only design.
MobilityGPT~\cite{haydari2026mobilitygpt} presents human mobility as an autoregressive generation task, applying a GPT decoder with road-connectivity constraints and reinforcement learning from trajectory feedback (RLTF) to produce semantically realistic urban paths.
MoveGPT~\cite{han2025movegpt} scales this approach to billion-scale datasets via a Spatially-Aware Mixture-of-Experts (SAMoE) Transformer trained across 16 cities, achieving state-of-the-art performance on a wide range of downstream mobility tasks.
MoveFM-R~\cite{meng2026movefmr} further adds large language model (LLM) semantic reasoning to foundation models for counterfactual trajectory simulation.
These models target urban mobility, but all require semantic zone IDs, road-network graphs, or absolute spatial embeddings: representations that are not part of the industrial setting considered here.

\subsection{Universal and Transferable Vehicle Trajectory Models}

TrajFM~\cite{lin2024trajfm} introduces a vehicle trajectory foundation model using a Spatio-Temporal Rotary (STR) position embedding module with a modality- and sub-trajectory-masking scheme spanning spatial, temporal, and point-of-interest (POI) channels, supporting region and task transferability.
UniTraj (Zhu et al.)~\cite{zhu2025unitraj} constructs the WorldTrace dataset of 2.45 million trajectories across 70 countries, and proposes Adaptive Trajectory Resampling and Self-supervised Trajectory Masking to handle heterogeneous sampling rates and quality.
UniTraj (Feng et al.)~\cite{feng2025unitraj} provides a benchmarking framework and documents that model performance drops sharply under cross-domain transfer.
TrajTok~\cite{xiong2026trajtok} proposes density-adaptive hexagonal spatial tokenisation combined with a factorised transformer encoder that applies separate self-attention to geometric and kinematic modalities before cross-attention fusion.
This keeps the geometric and kinematic representations distinct; IKTM instead \emph{discards} coordinates entirely.
BLUE~\cite{zhou2025blue} proposes a grid-free approach by truncating GPS decimal precision to form hierarchical patches, providing a coordinate-reduction baseline, though it still processes spatial information, unlike IKTM.

\subsection{Generative Trajectory Models}

TrajGDM~\cite{chu2023trajgdm} proposes a diffusion model for trajectory generation, treating the generation process as step-by-step uncertainty reduction and reporting at least 50\% Jensen--Shannon divergence (JSD) improvement over baselines on T-Drive and GeoLife.
Our autoregressive causal transformer provides direct temporal control and conditioning without the iterative denoising used by diffusion models at inference.
Dai et al.~\cite{dai2025trajfmad} survey large foundation models for trajectory prediction in autonomous driving, focusing on trajectory-language mapping, multimodal fusion, and constraint-based reasoning within the fully mapped, road-constrained autonomous driving context.

\subsection{Spatiotemporal Foundation Models: Surveys and Frameworks}

Several recent surveys place our contribution in context.
Fang et al.~\cite{fang2026stfmpipeline} review spatiotemporal foundation models through a pipeline lens (data preprocessing, embedding, pre-training, and adaptation), directly motivating our structured preprocessing pipeline.
Liang et al.~\cite{liang2025spatiotemporal} survey the broader landscape of spatiotemporal foundation model architectures, and Jin et al.~\cite{jin2026temporalmodels} extend this to time-series and spatiotemporal large models.
Chen et al.~\cite{chen2025geospatialssl} survey self-supervised learning for geospatial objects (points, polylines, polygons), providing context for our causal next-step training objective as an alternative to masked autoencoding.
Xu et al.~\cite{xu2025trajllm} survey how LLMs and autoregressive transformers are being integrated into physical trajectory prediction.
Nie et al.~\cite{nie2025llm4tr} survey LLM roles across transportation systems, while Zhang et al.~\cite{zhang2026llmmobility} specifically survey the use of LLMs for time-series forecasting and mobility analysis.

\subsection{Urban Foundation Models}

UrbanGPT~\cite{li2024urbangpt} connects spatial dependency encoders with instruction-tuned LLMs to predict urban traffic flows in zero-shot scenarios, but targets aggregated regional measurements rather than high-frequency individual kinematic sequences.
Li et al.~\cite{li2025urbanllm} survey the use of LLMs across urban computing tasks, and Zhang et al.~\cite{zhang2026ugi} survey urban foundation models and outline the landscape of urban general intelligence.
Choudhury et al.~\cite{choudhury2024towards} motivate trajectory foundation models through their broad societal applications; our coordinate-free design targets a complementary industrial setting where absolute-coordinate and map-based representations do not apply.

\subsection{Industrial Vehicle Mobility Models}
\label{sec:related_industrial}

Mobility simulation for industrial vehicles has traditionally relied on stochastic movement models derived from real telematics.
Our previous work~\cite{Amiri2024Model} extracts speed-change and direction-change statistics, each conditioned on the current speed class (four classes: 1--19, 20--39, 40--59, and 60+~km/h), from over six million telematics events across four industrial sites, and fits them with exponential curves ($R^2 = 0.8644$--$0.9863$).
The approach is coordinate-free and captures speed-conditioned turning statistics, but the speed-class state provides only single-step memory, direction-change magnitudes are estimated from 10-degree bins, and there is no trip-termination signal.
IKTM addresses these gaps through deep sequence modelling, continuous mixture heads, and duration conditioning.

\subsection{Positioning IKTM}
\label{sec:related_positioning}

Table~\ref{tab:comparison} summarises IKTM against representative methods.
Among prior work, the Markovian model described above is the closest in domain (industrial, coordinate-free), using single-step memory and no trip-termination mechanism.
Unlike the representative prior models surveyed in Table~\ref{tab:comparison}, IKTM combines (i) coordinate-free relative kinematics, (ii) probabilistic mixture heads for varied open-loop trajectories, (iii) duration conditioning for controllable termination, and (iv) zero-shot cross-site evaluation on real industrial telematics.

\begin{table}[!t]
\caption{Qualitative Comparison of Trajectory Generation Models}
\label{tab:comparison}
\centering
\scriptsize
\begin{threeparttable}
\begin{tabular}{lcccc}
\toprule
Model & Domain & Coord.-free & Prob.\ output & Industrial \\
\midrule
MobilityGPT~\cite{haydari2026mobilitygpt}     & Urban/Human   & \texttimes & \texttimes & \texttimes \\
MoveGPT~\cite{han2025movegpt}     & Urban/Human   & \texttimes & \texttimes & \texttimes \\
TrajFM~\cite{lin2024trajfm}         & Urban/Vehicle & \texttimes & \texttimes & \texttimes \\
UniTraj~\cite{zhu2025unitraj}        & Urban/Vehicle & \texttimes & \texttimes & \texttimes \\
TrajTok~\cite{xiong2026trajtok}        & Urban/Vehicle & \texttimes & \texttimes & \texttimes \\
TrajGDM~\cite{chu2023trajgdm}        & Urban/Human   & \texttimes & \checkmark & \texttimes \\
Amiri et al.~\cite{Amiri2024Model}  & Industrial    & \checkmark & \checkmark\tnote{*} & \checkmark \\
\textbf{IKTM (ours)} & \textbf{Industrial}  & \checkmark & \checkmark & \checkmark \\
\bottomrule
\end{tabular}
\begin{tablenotes}
\small
\item Coord.-free: no absolute spatial reference of any form, whether coordinates, grid or hexagonal cells, or road-network tokens.
\item UniTraj here is Zhu et al.'s model~\cite{zhu2025unitraj}. Feng et al.'s identically named UniTraj~\cite{feng2025unitraj} is a benchmarking framework rather than a generator, so it is not a row of this table.
\item Prob.\ output: models a continuous probability distribution over real-valued kinematic state (e.g., mixture density or diffusion) rather than deterministic point estimates or next-token classification over a discrete vocabulary. MobilityGPT and MoveGPT sample discrete road-link/location tokens, not a continuous state distribution; TrajGDM's diffusion operates in a continuous latent space that is subsequently decoded to discrete locations, and is counted as probabilistic here on the strength of that continuous generative process.
    \item[*] Amiri et al.\ samples from fixed parametric (exponential) distributions fitted per discrete speed class; IKTM predicts continuous, learned mixture parameters conditioned on the full kinematic history. Section~\ref{sec:baselines} provides an empirical comparison by re-implementing Amiri et al.\ as a rollout generator and evaluating it under the turn-rate and trip-length JSD protocol of Table~\ref{tab:baselines}.
\end{tablenotes}
\end{threeparttable}
\end{table}

\section{Trajectory Preprocessing and Standardisation}
\label{sec:preprocessing}

To train a purely coordinate-free model, raw geodetic coordinates must be converted into local kinematics while preserving accurate turning dynamics.
The preprocessing pipeline is shown in Fig.~\ref{fig:pipeline}.

\begin{figure}[!t]
\centering
\setlength{\unitlength}{1cm}
\begin{picture}(8,1.8)
  \put(0.0,0.5){\framebox(1.4,0.8){\scriptsize Raw GPS}}
  \put(1.4,0.9){\vector(1,0){0.4}}
  \put(1.8,0.5){\framebox(1.6,0.8){\scriptsize Filter/Segment}}
  \put(3.4,0.9){\vector(1,0){0.4}}
  \put(3.8,0.5){\framebox(1.5,0.8){\scriptsize B\'{e}zier Interp.}}
  \put(5.3,0.9){\vector(1,0){0.4}}
  \put(5.7,0.5){\framebox(1.5,0.8){\scriptsize $(v,\Delta\theta)$ seq.}}
  \put(0.0,0.0){\makebox(8,0.4){\footnotesize Output: coordinate-free kinematic time series at 1~Hz}}
\end{picture}
\caption{IKTM trajectory preprocessing pipeline.}
\label{fig:pipeline}
\end{figure}

\subsection{Trip Extraction and Noise Filtering}

Raw telematics records are filtered on GPS fix quality.
Trips are segmented at stationary events and whenever the gap between consecutive fixes exceeds 5~min.
A candidate trip is confirmed only if the vehicle moves more than 15~m from its start point within its first four fixes, including the start event and up to three subsequent fixes. This criterion rejects stationary GPS drift; the candidate must also contain at least 5 checkpoints and accumulate at least 100~m of travelled distance.
For Site~A, this yields 180,198 candidate trips, reduced to 179,151 after the subsequent per-vehicle GPS-drift cleaning and out-of-scope filtering applied during time-series generation.
For each connection, the cleaner computes the maximum reachable distance from elapsed time, median-smoothed endpoint speeds, and per-vehicle speed/acceleration caps; the connection is infeasible when its chord, the straight line joining the two consecutive checkpoints, exceeds 1.25 times that budget plus a 5~m tolerance.
A trip is discarded rather than salvaged if cleaning removes more than 60\% of its checkpoints, and trips with head-to-tail distance above 5~km are outside this industrial-yard scope.

\subsection{Orientation and Origin Normalisation}

Geodetic coordinates $(\mathrm{lat}_i, \mathrm{lon}_i)$ are projected to local Cartesian coordinates $(x_i, y_i)$ relative to the trip origin.
All coordinates are then rotated so that the first movement segment aligns with the positive $x$-axis, ensuring the absolute departure bearing never enters the training signal.
A synthetic acceleration prefix (rest to the first recorded speed) and a synthetic deceleration suffix (final recorded speed back to rest) teach the model to start from and return to rest.
Each ramp's duration is derived from that boundary checkpoint's recorded speed and rounded up to the nearest whole second, sizing the ramp for a conservative $2$~m/s$^2$ acceleration.

\subsection{Heading-Aware Curved Interpolation}
\label{sec:bezier}

Industrial telematics is sparsely and irregularly sampled: the median inter-checkpoint gap is 5~s, with a 99th percentile of 45~s.
Standard linear interpolation means the vehicle travels in perfectly straight lines between checkpoints, concentrating all turning at checkpoint junctions.
On our Site~A dataset, linear interpolation produces 93.5\% of 1-second steps with a heading change of exactly zero, concentrating continuous turning at the checkpoints rather than across each segment.

To restore turns, we interpolate each segment $k \to k{+}1$ with a cubic B\'{e}zier curve:
\begin{equation}
  \mathbf{B}(t) = (1{-}t)^3 P_0 + 3(1{-}t)^2 t C_0 + 3(1{-}t) t^2 C_1 + t^3 P_1,
\end{equation}
where $P_0, P_1$ are the checkpoint positions and $C_0, C_1$ are control points placed along the recorded GPS headings at $P_0$ and $P_1$. Tangent magnitudes are set to one-third of the chord distance to limit geometric overshoot beyond the chord. This spatial-path constraint does not separately bound the resulting speed.
At nodes where the speed is below 1.0~m/s or the recorded heading points backward relative to the segment chord, the tangent falls back to the chord direction; query times beyond the last distinct checkpoint are clamped to the final distinct checkpoint.

This curved interpolation does three things.
First, it passes exactly through every kept checkpoint with a distinct position. Fixes with duplicate timestamps, or with the same position as the preceding fix, are dropped beforehand, as are the kinematically-infeasible GPS-drift ``flyaway'' checkpoints removed by the earlier drift-cleaning stage.
Second, it distributes turning continuously, reducing exact-zero $\Delta\theta$ steps from 93.5\% to 44.2\%.
Third, it preserves reconstruction accuracy: integrating the curve-derived position and speed reproduces the checkpoint reference with a median per-axis maximum error of 0.414~mm. Across all 179,151 Site~A trips, 99.97\% reconstruct to less than 1~cm of error (worst case 1.64~cm). Each trip contributes the largest of its $x$ and $y$ deviations over its retained checkpoints, so the figures are per-axis deviations rather than Euclidean distances, which they bound to within a factor of $\sqrt{2}$. This accuracy measure uses the curve's own path-derived position and speed, not the independently filtered speed published in the exported kinematics, which is decoupled from the reconstructed path by design.

Total per-trip absolute turning also rises substantially: from a median of 260.0\degree{} in the raw checkpoint-to-checkpoint record, computed over the same drift-cleaned checkpoints the interpolated series are built from, and 459.7\degree{} under linear interpolation, to 786.0\degree{} under the curved export.
The sparse raw record and linear interpolation both concentrate each segment's entry and exit curvature into a single instantaneous heading change, which the curved interpolant instead distributes across the segment.
The totals reflect three different heading representations. The raw baseline measures changes between recorded compass headings at irregular checkpoints (median spacing 5~s); the linear baseline samples the fixed bearing of each checkpoint-to-checkpoint chord at 1~Hz; and the curved export samples the changing bearing of a B\'{e}zier path at 1~Hz, with the recorded headings defining its endpoint tangents.
The restored turning is coherent rather than noisy. Across all Site~A trips, the sign-change rate between consecutive nonzero $\Delta\theta$ steps is 0.20, well below the 0.5 expected of uncorrelated jitter under a balanced (equal left/right) sign marginal.
A substantially imbalanced marginal would lower that null value somewhat, but not to anything near 0.20, so the qualitative conclusion (coherent turning, not jitter) is insensitive to this assumption.
Of the curved steps, 2.9\%, 1.3\%, and 0.5\% exceed 90, 120, and 150\degree{}/s, respectively. Only 3.0--5.3\% of that tail coincides with near-stationary speed ($<0.5$~m/s in the curve-derived speed used by this check, where GPS-derived heading is least reliable), so the extreme tail is not a near-stationary-noise artefact; it is in any case excluded from the headline metric by the turn-rate JSD's $45\degree$ cutoff (Section~\ref{sec:metrics}).
This approach is separate from prior adaptive resampling strategies such as UniTraj's Adaptive Trajectory Resampling~\cite{zhu2025unitraj}, which targets heterogeneous sampling rates rather than the turn-erasure problem addressed here.

\subsection{Kinematic Feature Representation}

For each second $t$, the standardised trajectory is represented as
\begin{equation}
    X_t = [v_t,\; \sin(\Delta\theta_t),\; \cos(\Delta\theta_t),\; \mathit{BOS}_t],
\end{equation}
where $v_t$ is the speed, normalised to zero mean and unit variance using training-set statistics; $\Delta\theta_t$ is the heading change in radians, whose sine/cosine decomposition handles the $-\pi/\pi$ wrap-around; and $\mathit{BOS}_t$ is a beginning-of-sequence (BOS) flag.
With duration conditioning enabled (Section~\ref{sec:duration}), a normalised remaining-time feature is appended, giving a 5-dimensional input.

\section{IKTM System Architecture}
\label{sec:architecture}

IKTM is a causal, decoder-only transformer~\cite{vaswani2017attention} that predicts the distribution of the next kinematic step from the history of past steps.

\subsection{Causal Transformer Backbone}

Input features $X_t$ are projected into a latent space of dimension $d_{\text{model}} = 128$ and combined with learned positional encodings.
The backbone stacks four self-attention layers with four heads each and a feed-forward dimension of 512 (dropout 0.1); the full model (backbone plus prediction heads) totals ${\approx}0.8$M parameters.
Causal attention masks ensure that the prediction for step $t{+}1$ depends only on steps $0, \ldots, t$.
Following the standard convention for GPT-style architectures~\cite{merten2025trajfromscratch}, the backbone is a self-attention-only stack implemented with PyTorch's \texttt{TransformerEncoder}, an explicit causal mask, and no cross-attention. PyTorch's \texttt{TransformerDecoder} includes a cross-attention sublayer for encoder-conditioned sequence-to-sequence models, which is unused here. Accordingly, ``decoder-only'' refers to the architecture family rather than the literal PyTorch class name.
The maximum attention context is 60 tokens (60 seconds at 1~Hz), a design choice balancing recent-history coverage against the quadratic cost of a longer window.

\subsection{Probabilistic Mixture Heads}
\label{sec:prob_heads}

Deterministic regression (e.g., mean squared error (MSE) or Huber loss) drives the head toward a point estimate of the conditional mean.
During open-loop rollouts, this causes the model to converge to a fixed point: the predicted mean values for speed and heading change at each step are fed back as the next input. The next prediction again lies near those means, producing a constant-speed, constant-curvature trajectory that never stops.
We address this behaviour by replacing point-estimate heads with mixture distributions and sampling during rollout, in line with the generative approach of TrajGDM~\cite{chu2023trajgdm} for diffusion trajectories.

\textbf{Speed head}: A GMM with $K_v = 3$ components:
\begin{equation}
    p(v_{t+1} \mid X_{:t}) = \sum_{j=1}^{K_v} \pi_j \,\mathcal{N}(v_{t+1};\, \mu_j,\, \sigma_j),
\end{equation}
where the mixture weights $\pi_j$, means $\mu_j$, and scales $\sigma_j$ (standard deviations, not variances) are produced by a linear head on the transformer output (weights via softmax, scales via softplus to enforce positivity; their dependence on the history $X_{:t}$ is suppressed for readability).
The mixture is defined over \emph{normalised} speed, using the same zero-mean/unit-variance scale as the input features. Sampled values are de-normalised to m/s and clamped at generation time, as described below, before being emitted or fed back.

\textbf{Heading head}: A mixture of von Mises distributions with $K_\theta = 5$ components to handle the circularity of $\Delta\theta$:
\begin{equation}
    p(\Delta\theta_{t+1} \mid X_{:t}) = \sum_{m=1}^{K_\theta} \omega_m \,\mathcal{M}(\Delta\theta_{t+1};\, \mu_m,\, \kappa_m),
\end{equation}
where $\mu_m$ is the mean direction and $\kappa_m$ the concentration parameter.
This factorised kinematic parameterisation is conceptually aligned with TrajTok's factorised encoder~\cite{xiong2026trajtok}, which separates geometric and kinematic modalities in early attention layers; IKTM discards the geometric modality entirely.

\textbf{Stop head}: A sigmoid classifier outputs the stop probability $p_{\text{stop}}$, trained with class-balanced binary cross-entropy (BCE) to counteract the rarity of stop events (a single positive label at the final timestep of each trip).

During generation, we sample from the mixture heads with a temperature parameter $T$: component logits and von Mises concentrations are divided by $T$, and Gaussian standard deviations are multiplied by $T$, so $T < 1$ sharpens all sampled distributions.
Sampling injects some diversity, but Section~\ref{sec:rollout_quality}'s duration-free temperature sweep shows this alone does not secure reliable termination at any tested temperature: under the same 1,250-sample cap that the duration-conditioned variants use, the natural stop rate reaches 78\% at $T=0.9$ and is 0\% at $T \le 0.5$, including the near-greedy $T=0.01$ setting.
At the selected default of $T=0.2$, the duration-free variant does not self-terminate under this sweep, so reliable termination at that operating point comes from duration conditioning rather than from sampling diversity.
To suppress rare, physically implausible tail samples, generated speeds are clamped to $[0,\, 13.9]$~m/s; negative Gaussian samples are forced to zero. The upper bound is the measured 99th-percentile speed over Site~A's train and validation devices, with the test split excluded (13.889~m/s; raw maximum 26.944~m/s). The clamped value is re-normalised and fed back as the next-step input state. Thus, the autoregressive history never exceeds the generation-time ceiling, while training consumes the unclamped speed series.

\subsection{Duration Conditioning}
\label{sec:duration}

To enable controllable, reliable termination, we expand the input feature vector with a normalised remaining-time feature:
\begin{equation}
    \tau_t = \frac{\min(\max(D - t,\, 0),\, 300~\text{s})}{\sigma_D},
\end{equation}
where $D$ is the target trip duration in elapsed seconds, excluding only the initial rest/BOS token (row 0), and $\sigma_D = 60$~s is a scaling factor; the synthetic acceleration/deceleration ramps (Section~\ref{sec:preprocessing}) are counted like any other exported second.
During training, $D$ is set to the true trip length, so the model learns to associate a decrementing remaining-time signal with deceleration and stop behaviour.

For generation, a target duration $D$ is sampled from a \textbf{learned log-duration prior},
\begin{equation}
    \log(D) \sim \mathcal{N}(\mu_{\log D},\, \sigma_{\log D}^2),
\end{equation}
whose parameters $(\mu_{\log D}, \sigma_{\log D})$ are learned jointly with the rest of the model by minimising the log-duration negative log-likelihood (NLL). This NLL is evaluated at each trip's BOS window only, so trips are not overweighted by their number of overlapping sliding windows, and on both the plain and sign-mirrored copy of that window (Section~\ref{sec:preprocessing}).
The mirroring duplicates every trip's contribution equally rather than reweighting any trip relative to another, since the NLL depends only on trip length, not on the mirrored heading sign. Samples are clipped to $[2, 1000]$~s.
For the published checkpoint the prior fits a median trip duration of 79.9~s ($\exp(\mu_{\log D})$) with a log-duration standard deviation of 0.782 (obtained from $\sigma_{\log D} = \mathrm{softplus}(\log\sigma_{\log D}) + 10^{-2}$), so the fitted prior over $D$ is lognormal with those parameters, before the rounding to whole seconds and clipping described above.
Because we do not observe destination or route intent, the coordinate-free kinematic dynamics alone provide no signal about intended trip length, so the marginal log-duration distribution serves as the natural prior.
At each generation step, $\tau_t$ is decremented, guiding the model to wind down its speed and emit the stop signal when the countdown expires; the clamp at zero binds only if a rollout overshoots its target.
The upper clip at 300~s binds in the other direction: because the prior samples targets from $[2, 1000]$~s, a target beyond 300~s leaves $\tau_t$ pinned at its ceiling for the opening $D - 300$ seconds, so over that interval the feature indicates only that the end is distant, without resolving how distant. Section~\ref{sec:rollout_quality} reports length tracking in this regime.

\subsection{Training Objective and Setup}
\label{sec:training_setup}

The full model is trained by minimising
\begin{equation}
    \mathcal{L} = \mathcal{L}^{v}_{\text{NLL}} + \mathcal{L}^{\Delta\theta}_{\text{NLL}} + \lambda\, \mathcal{L}^{\text{stop}}_{\text{BCE}} + \mathcal{L}^{D}_{\text{NLL}},
\end{equation}
the sum of the speed and heading mixture negative log-likelihoods, the class-balanced stop BCE ($\lambda = 1$; positive-class weight computed automatically from the label imbalance, ${\approx}99.6$ on Site~A), and the log-duration prior NLL.
The deterministic variants replace the speed and heading mixture NLLs with Huber regression losses on $[v, \sin\Delta\theta, \cos\Delta\theta]$; duration-conditioned variants retain the log-duration prior NLL.
All variants are trained on sliding windows of up to 60 steps (50\% overlap) with AdamW (learning rate $10^{-3}$, weight decay $10^{-4}$), cosine annealing, and batch size 128 on a cloud NVIDIA RTX~4090. The checkpoint with the best validation loss is kept. All variants, including the deterministic and probabilistic-only ablation baselines, train for 30 epochs, matching the full IKTM (probabilistic heads with duration conditioning). Validation loss is noisy and non-monotonic; the kept full-IKTM checkpoint is from epoch 24 of 30. Every training-split window is duplicated with its heading sign flipped ($\sin\Delta\theta \to -\sin\Delta\theta$), making the training distribution exactly left/right symmetric. Validation windows and all real-data evaluation references are left unmirrored.

\section{Experimental Evaluation}
\label{sec:evaluation}

\subsection{Dataset and Cross-Site Split}

We evaluate on real telematics logs from four industrial sites, anonymised as Site~A--D for commercial confidentiality: Site~A (179,151 trips; 19.6~million 1~Hz samples), Site~B (13,749 trips), Site~C (21,503 trips), and Site~D (23,844 trips).
Training, validation, and testing use Site~A only, under a 70/15/15 device-level split (hard-coded seed 42 on its own RNG stream, numerically coincident with the canonical evaluation seed introduced below but unrelated to it): no device contributes data to more than one of the three sets, preventing per-device leakage. Checkpoint selection (best validation loss) and sampling-temperature selection (Section~\ref{sec:rollout_quality}) are both performed on the validation split; all headline Site~A next-step and rollout metrics reported below are computed on the untouched test split.
Training itself, including weight initialisation and batch shuffling, is not seeded. The published checkpoint, rather than a retraining command, is therefore the reproducible source behind the reported numbers. Evaluation-time rollout and duration-prior sampling use a separate seed, 42 by default, varied over $\{1,\ldots,100\}$ in the seed-variance check (Section~\ref{sec:rollout_quality}). Training ran on a cloud NVIDIA RTX~4090 (CUDA), while all reported evaluations, including the 100-seed sweep, ran locally on Apple Silicon (Metal Performance Shaders, MPS). Seeded rollout sampling reproduces exactly within a given hardware backend and generator mode, sequential or batched (Section~\ref{sec:rollout_quality}).
Site~B, Site~C, and Site~D are held out entirely as zero-shot OOD test sets.
This train-on-one-site, test-zero-shot protocol follows the cross-domain evaluation philosophy of UniTraj (Feng et al.)~\cite{feng2025unitraj}, which documented sharp cross-dataset performance degradation for existing trajectory predictors under domain shift.
Table~\ref{tab:dataset_stats} reports per-site device, trip, and sample counts alongside speed/duration/turn-rate percentiles and device-level fleet composition, so that the zero-shot transfer results below can be read against how each OOD site's operating conditions and vehicle mix compare to Site~A's, rather than just against its trip count.
The four fleets differ in composition as well as in scale: the training site's fast-tier share is 1.8\% of devices, against 10.0--16.5\% at the held-out sites.

\begin{table}[!t]
\caption{Per-Site Dataset Statistics and Fleet Composition}
\label{tab:dataset_stats}
\centering
\scriptsize
\begin{threeparttable}
\begin{tabular}{lcccc}
\toprule
 & Site~A & Site~B & Site~C & Site~D \\
\midrule
Devices              & 450        & 130       & 164       & 125 \\
Trips                & 179,151    & 13,749    & 21,503    & 23,844 \\
1~Hz samples         & 19,600,183 & 1,720,760 & 2,547,585 & 2,204,640 \\
Speed p95 (m/s)      & 10.00      & 6.94      & 11.72     & 8.89 \\
Duration p95 (s)     & 311        & 439       & 380       & 251 \\
Turn-rate p95 (deg/s) & 50.37     & 41.32     & 47.79     & 56.50 \\
\midrule
\multicolumn{5}{l}{\emph{Fleet composition (\% of devices, by speed tier)}} \\
Slow ($<$12~m/s)     & 74.4       & 70.8      & 65.2      & 65.6 \\
Medium ($<$20~m/s)   & 22.2       & 9.2       & 12.8      & 19.2 \\
Fast                 & 1.8        & 10.0      & 16.5      & 10.4 \\
Unclassified         & 1.6        & 10.0      & 5.5       & 4.8 \\
\bottomrule
\end{tabular}
\begin{tablenotes}
\small
\item Duration: elapsed seconds per trip (row count $-1$). Turn-rate: $|\Delta\theta|$ per second. All percentiles computed directly from the published \texttt{data/timeseries} export.
\item Speed tier is assigned per device from its p99 speed (\texttt{classify\_vehicles.py}); ``unclassified'' is that stage's fallback for devices with too few speed samples for a stable percentile. Shares are over devices, so they describe fleet make-up rather than utilisation. The training site has the smallest fast-tier share of the four (1.8\% of devices, against 10.0--16.5\% at the held-out sites); Section~\ref{sec:robustness} relates this to the speed-marginal results.
\end{tablenotes}
\end{threeparttable}
\end{table}

\subsection{Evaluation Metrics}
\label{sec:metrics}

\textbf{Next-step prediction metrics} (teacher-forced):
\begin{itemize}
    \item \textit{Speed RMSE} (m/s): root mean squared error of predicted vs.\ actual speed at each step, taking the mixture mean as the prediction for probabilistic variants.
    \item \textit{Heading RMSE} (degrees): RMSE of $\Delta\theta$, evaluated on the mixture's point estimate, the weight-weighted circular mean of its component means.
    \item \textit{Stop BCE}: unweighted binary cross-entropy of the stop classifier, averaged per token (training uses a class-balanced BCE with an automatic positive-class weight, $\approx$99.6 on Site~A, Section~\ref{sec:training_setup}, so this evaluation metric is not directly comparable to the training loss). Because stop events are rare ($\approx$1\% of tokens), a near-zero unweighted BCE reflects easy majority-class agreement more than positive-class quality on its own, and should be read alongside the precision/recall figures below.
    \item \textit{Stop precision / recall}: precision and recall of stop-event detection at a 0.5 threshold.
\end{itemize}

\textbf{Rollout quality metrics} are computed autoregressively from BOS with seed 42 throughout. Rollouts are capped at 1,250 samples, including the initial BOS/rest sample (1,249 elapsed seconds under the elapsed-time convention defined below), which lies above the support of both duration-target \emph{arms}: the model's learned prior (\emph{prior-length}) and each held-out site's own empirical trip-length histogram (\emph{oracle-length}).
\begin{itemize}
    \item \textit{Natural stop rate}: fraction of rollouts terminated by the stop head ($p_{\text{stop}} > 0.5$) before the cap.
    \item \textit{Length-tracking error}: generated trip length minus sampled target duration, both measured in elapsed seconds excluding the initial BOS/rest sample ($\mu \pm \sigma$ over $N$ rollouts; $\sigma$ is the population standard deviation).
    \item \textit{Turn-rate JSD}: Jensen--Shannon divergence (base-2, in bits; range $[0,1]$) between 50-bin histograms of per-step $|\Delta\theta|$ over $[0\degree, 45\degree]$ from generated rollouts and from real data. The real reference is drawn from the published, heading-aware-interpolated export used throughout this paper, not the raw sparse GPS record, so it inherits the curved-interpolation choice described in Section~\ref{sec:preprocessing}. A minority tail exceeds $45\degree$ (p99${}=126.7\degree$ over the Site~A curved export) and is excluded from the histogram for both real and generated data. This excluded tail is 4.5--6.8\% of real steps but only 0.9--1.0\% of generated steps across the four sites under the oracle-length arm, and 0.8--1.1\% under the prior-length arm (canonical seed, Table~\ref{tab:rollout} settings), so generated rollouts contain fewer large turns relative to real data even though the retained $[0\degree,45\degree]$ shape matches closely (Table~\ref{tab:rollout}); the metric therefore chiefly captures small-turn shape. Both histograms are renormalised to sum to 1 over the retained range before the JSD is computed ($p,q \gets p/\!\sum p,\, q/\!\sum q$), so the comparison itself is unaffected by the differing exclusion rates. Lower is better; 0 indicates identical distributions.
    \item \textit{Trip-length JSD}: the same divergence between 50-bin trip-length histograms over $[0, 1200]$~s. Trips longer than 1,200~s fall outside the histogram for both real and generated data, but no site has more than 0.023\% of trips there and the per-site p95 is at most 439~s. The duration prior clips sampled targets to $[2, 1000]$~s (Section~\ref{sec:duration}), so wherever targets come from that prior (Site~A under either arm, and every site under the prior-length arm), generated lengths cannot reach the $(1000, 1200]$~s tail; the oracle-length arm instead draws Sites~B--D's targets from each site's own empirical histogram, which the prior's 1,000~s clip does not bound, though it shares the $[0, 1200]$~s grid above.
\end{itemize}

\subsection{Ablation Results on Site A (In-Distribution)}

Table~\ref{tab:results_site_a} reports next-step metrics on the Site~A held-out test split for each model variant, spanning the full $2\times2$ heads$\times$duration-conditioning grid.

\begin{table}[!t]
\caption{Next-Step Prediction Metrics: Site A Test Split (In-Distribution)}
\label{tab:results_site_a}
\centering
\scriptsize
\begin{threeparttable}
\begin{tabular}{lcccc}
\toprule
Model & Speed & Heading & Stop & Stop \\
      & RMSE  & RMSE   & BCE  & Prec / Rec \\
      & (m/s) & (deg)  &      & \\
\midrule
Det.\ Transformer    & 0.16 & 21.6\degree & 0.0017 & 0.953 / 0.999 \\
Prob.\ Transformer   & 0.15 & 22.7\degree & 0.0036 & 0.913 / 0.999 \\
Det.+Dur.\ Transformer & 0.13 & \textbf{19.8}\degree & $0.0000^{\dagger}$ & $\mathbf{1.000 / 1.000}^\dagger$ \\
\textbf{IKTM} (Prob.+Dur.) & \textbf{0.12} & 21.3\degree & $0.0003^{\dagger}$ & $0.992 / 1.000^\dagger$ \\
\bottomrule
\end{tabular}
\begin{tablenotes}
\small
\item Det.: deterministic regression heads. Prob.: probabilistic mixture heads.
\item Dur.: duration conditioning. For probabilistic variants, RMSE is from the mixture point estimate. Stop BCE is rounded to 4 decimal places; ``0.0000'' denotes $<5\times10^{-5}$, not an exact zero.
\item Prob.\ Transformer's teacher-forced metrics are temperature-independent (computed from the mixture point estimate, not a sampled rollout); its reproduction command uses $T=0.2$ for its qualitative BOS demo, matching the selection swept for the full IKTM (Section~\ref{sec:rollout_quality}).
\item $^\dagger$Oracle-$\tau$: unlike Det./Prob., the duration-conditioned rows' teacher-forced stop metrics (BCE and Prec/Rec) are conditioned on the true remaining time $\tau$, which is unavailable at generation time. Their speed and heading metrics are likewise conditioned on $\tau$ because all three heads share the same conditioned input. See the direct rollout-based termination results (Section~\ref{sec:rollout_quality}) and the $\tau$-control ablation (Section~\ref{sec:tau_control}) for the corresponding evidence.
\end{tablenotes}
\end{threeparttable}
\end{table}

Heading RMSE is closely comparable across variants (19.8--22.7\degree), reflecting the similar teacher-forced next-step prediction task; speed RMSE varies more across variants (0.12--0.16~m/s).
Each row is one training run and training is not seeded (Section~\ref{sec:training_setup}), so the sub-0.05~m/s speed-RMSE and sub-3\degree{} heading-RMSE differences between rows are not separated from run-to-run variation.
The largest numerical effect of duration conditioning is in the stop head, where it reduces stop BCE from 0.0017--0.0036 (Det./Prob.) to 0.0000--0.0003 and raises stop precision from 0.91--0.95 to 0.99--1.00.
Det.+Dur.\ reaches teacher-forced stop precision and recall of 1.000 for both measures and the lowest heading RMSE of the four variants, so use of the countdown does not depend on probabilistic mixture heads.
This teacher-forced comparison gives the duration-conditioned variants additional information. The remaining-time feature is computed from the true trip end (Section~\ref{sec:duration}) and fed to the shared backbone at every step. It provides both duration-conditioned variants with an almost-direct stop-label signal that the Det.\ and Prob.\ baselines do not receive, together with weaker speed and heading signals because all three heads share the input. The teacher-forced numbers show that the model exploits this signal when supplied; they do not show that it anticipates termination from kinematics alone. Section~\ref{sec:tau_control} quantifies the oracle countdown's contribution, Section~\ref{sec:rollout_quality} evaluates self-termination with sampled targets, and the cross-site speed and heading results in Section~\ref{sec:ood} carry the same oracle-$\tau$ qualification.

The heading RMSE of 19.8--22.7\degree{} should be interpreted relative to the naive baseline of predicting zero heading change, which yields an RMSE equal to the root-mean-square $\Delta\theta$ (25.56\degree{} on the Site~A test split, excluding each trip's BOS row, which the next-step task never predicts).
Computed from unrounded RMSEs against the 25.56\degree{} baseline, our models reduce heading RMSE by 11.2--22.4\% relative to it (Det.\ 15.3\%, Prob.\ 11.2\%, Det.+Dur.\ 22.4\%, IKTM 16.5\%), confirming that the B\'{e}zier-interpolated training data contains a learnable turning signal.
Speed RMSE has the same kind of naive reference, on the same populations. A persistence predictor ($v_{t+1} = v_t$, that is, carry the current speed forward one step) scores 0.3073~m/s on the Site~A test split and 0.2643, 0.2951, and 0.3306~m/s at Sites~B, C, and D, respectively.
IKTM reduces those errors by 61.7\%, 51.5\%, 52.1\%, and 54.4\%, respectively, a wider margin than it achieves on heading, so the absolute speed scale is not simply being carried forward from the previous step.
Both reductions are computed from duration-conditioned rows, which are oracle-$\tau$ (Table~\ref{tab:results_site_a}), while the naive baselines receive no such signal. At Site~A, where the $\tau$-control of Section~\ref{sec:tau_control} was run, IKTM's speed reduction over persistence is 34.7--37.1\% under the two controls, and its heading reduction over the zero-change baseline is 10.4--11.1\%; the Det.\ and Prob.\ heading reductions (15.3\% and 11.2\%) carry no such conditioning.

\subsection{$\tau$-Control Ablation: How Much Is the Oracle Countdown?}
\label{sec:tau_control}

Table~\ref{tab:results_site_a}'s duration-conditioned rows are teacher-forced with the true remaining time $\tau$ at every step.
To quantify how much of their speed, heading and stop advantage is specifically this oracle countdown rather than learned kinematics, we re-evaluate the full IKTM with $\tau$ replaced by two controls, holding checkpoint, test split and teacher forcing fixed.
The \emph{prior} control samples one value per window from the model's own learned duration prior (Section~\ref{sec:duration}) and holds it constant across that window's timesteps, giving a plausible but mismatched signal that does not count down.
The \emph{none} control is a constant ``always far from the end'' value, the duration-clip ceiling, for every timestep of every window.
Table~\ref{tab:tau_control} reports both controls against the oracle row.

\begin{table}[!t]
\caption{$\tau$-Control Ablation: IKTM, Site~A Test Split, Teacher-Forced}
\label{tab:tau_control}
\centering
\scriptsize
\begin{tabular}{lcccc}
\toprule
$\tau$ source & Speed & Heading & Stop & Stop \\
              & RMSE  & RMSE   & BCE  & Prec / Rec \\
              & (m/s) & (deg)  &      & \\
\midrule
Oracle (true $\tau$) & 0.12 & 21.3\degree & 0.0003 & 0.992 / 1.000 \\
Prior (fixed, const.) & 0.20 & 22.9\degree & 0.0624 & 0.685 / 0.005 \\
None (ceiling, const.) & 0.19 & 22.7\degree & 0.0690 & 0.963 / 0.002 \\
\bottomrule
\end{tabular}
\end{table}

Stop recall falls from 1.000 under the true countdown to 0.005 (prior) and 0.002 (none). Thus, the near-perfect teacher-forced stop metrics in Table~\ref{tab:results_site_a} arise almost entirely from the oracle signal rather than anticipation of trip end from kinematics alone. Speed RMSE increases from 0.12 to 0.19--0.20~m/s (64--70\%), while heading RMSE increases from 21.3\degree{} to 22.7--22.9\degree{} (${\approx}$7\%). These changes are consistent with the shared backbone: all three heads receive the same conditioned input, so an incorrect $\tau$ also affects speed and heading. Generation does not receive oracle $\tau$.
Section~\ref{sec:rollout_quality} instead samples termination targets from IKTM's learned prior and reports a 100\% natural stop rate (Table~\ref{tab:rollout}).
Duration conditioning therefore acts on termination, while next-step manoeuvre accuracy is comparable across all four variants (0.12--0.16~m/s speed, 19.8--22.7\degree{} heading).

\subsection{Zero-Shot OOD Generalisation}
\label{sec:ood}

Table~\ref{tab:results_ood} shows zero-shot next-step performance of the full IKTM across all four sites.
The deterministic and probabilistic variants follow a qualitatively similar pattern, though the probabilistic-only variant's speed RMSE rises from 0.15~m/s in-distribution to 0.19--0.23~m/s OOD, against the roughly flat 0.15--0.18~m/s cross-site band of the deterministic variant.

\begin{table}[!t]
\caption{Cross-Site Next-Step Metrics: IKTM (Site~A In-Distribution (ID), Sites~B--D Zero-Shot OOD)}
\label{tab:results_ood}
\centering
\scriptsize
\begin{threeparttable}
\begin{tabular}{lcccc}
\toprule
Site & Speed & Heading & Stop & Stop \\
     & RMSE  & RMSE   & BCE  & Prec / Rec \\
     & (m/s) & (deg)  &      & \\
\midrule
Site~A (ID-test) & 0.12 & 21.3\degree & 0.0003 & 0.992 / 1.000 \\
Site~B (OOD)     & 0.13 & 19.6\degree & 0.0002 & 0.992 / 1.000 \\
Site~C (OOD)     & 0.14 & 20.7\degree & 0.0002 & 0.992 / 1.000 \\
Site~D (OOD)     & 0.15 & 23.5\degree & 0.0003 & 0.992 / 1.000 \\
\bottomrule
\end{tabular}
\begin{tablenotes}
\small
\item Model trained exclusively on Site~A; Site~B, Site~C, Site~D never seen during training.
\item Speed RMSE is quoted to two decimal places throughout this table. Table~\ref{tab:loo_matrix} reports its cells to three, so Site~A's $0.12$~m/s here and the $0.118$~m/s home cell there are the same measurement at two precisions, not two figures.
\item Stop recall is 1.000 across all sites; as in Table~\ref{tab:results_site_a}, all of IKTM's teacher-forced metrics here (speed, heading, and stop) are oracle-$\tau$ (conditioned on the true remaining time, fed to the shared backbone at every step), and the stop numbers specifically should be read alongside the direct rollout-based termination results (Section~\ref{sec:rollout_quality}) and the $\tau$-control ablation (Section~\ref{sec:tau_control}).
\end{tablenotes}
\end{threeparttable}
\end{table}

In the teacher-forced next-step setting, both the kinematic turning structure and the absolute speed scale transfer without fine-tuning: heading RMSE spans 19.6--23.5\degree{} across sites, and speed RMSE is 0.13--0.15~m/s OOD against 0.12~m/s in-distribution. Section~\ref{sec:robustness} reports the corresponding unconditional quantity, the speed marginal of a free-running rollout.
A leave-one-site-out check tests whether this pattern depends on Site~A being the home site. A separate IKTM is trained on each of Sites~B, C, and D in turn and evaluated zero-shot on the other three. Table~\ref{tab:loo_matrix} reports the full $4{\times}4$ home-vs-cross-site speed-RMSE matrix; home cells are each checkpoint's own held-out split rather than a uniform ``test'' set, as the tablenote explains.
Values stay within a narrow 0.118--0.165~m/s band regardless of which site is home, with no systematic home-vs-OOD gap: each home cell sits within 0.022~m/s of the mean of its own row's cross-site cells, and is the \emph{higher} of the two in two of the four rows.
Stop recall remains 1.000 across all OOD sites under teacher forcing with the true countdown supplied (oracle-$\tau$); the sampled-duration rollout results below give the direct evidence that termination is reliable without access to the true remaining time.

\begin{table}[!t]
\caption{Leave-One-Site-Out Speed RMSE (m/s), $4\times4$ Home-vs-Cross-Site Matrix}
\label{tab:loo_matrix}
\centering
\scriptsize
\begin{threeparttable}
\begin{tabular}{lcccc}
\toprule
Train $\downarrow$ / Test $\rightarrow$ & Site~A & Site~B & Site~C & Site~D \\
\midrule
Site~A              & 0.118$^*$ & 0.128     & 0.141     & 0.151 \\
Site~B              & 0.133     & 0.140$^*$ & 0.147     & 0.150 \\
Site~C              & 0.130     & 0.119     & 0.134$^*$ & 0.144 \\
Site~D              & 0.125     & 0.148     & 0.165     & 0.163$^*$ \\
\bottomrule
\end{tabular}
\begin{tablenotes}
\small
\item[$*$] Home cell (train site = test site). The Site~A home cell is measured on the production checkpoint's held-out \emph{test} split (Section~\ref{sec:training_setup}); the Site~B/C/D home cells are measured on each leave-one-site-out checkpoint's held-out \emph{validation} split (much smaller subsets than the OOD cells' full datasets), since those checkpoints were trained only for this cross-site check and have no separate test split of their own. They also predate the locked 70/15/15 split and were trained under an earlier 80/20 train/validation partition, so the matrix compares runs at two training-set sizes. The three home validation splits are unequal in size: 2,174 trips at Site~B against 4,321 at Site~C and 4,390 at Site~D. Full range across all 16 cells: 0.118--0.165~m/s. Heading RMSE (same runs) ranges 19.6--26.4\degree{} across all cells.
\end{tablenotes}
\end{threeparttable}
\end{table}

For comparison, UniTraj (Feng et al.)~\cite{feng2025unitraj} reports sharp degradation of coordinate-based trajectory predictors under cross-dataset domain shift.
Next-step kinematics therefore transfer without fine-tuning, and do not depend on Site~A being the home site.

\subsection{Rollout Quality}
\label{sec:rollout_quality}

Unlike coordinate-based generators, which can terminate a rollout via spatial proximity to a destination, a coordinate-free model has no such signal and must learn termination purely from kinematic and duration cues; the diagnostics below evaluate whether this succeeds in open-loop generation, independent of the teacher-forced stop-head metrics reported above.

With deterministic regression heads and no duration conditioning, the autoregressive BOS rollout converges to a fixed point (observed on the seeded diagnostic rollout, $N{=}1$): after an initial transient, speed saturates at the 13.9~m/s generation ceiling ($13.90 \pm 0.00$~m/s) at a near-constant turn rate ($-0.616 \pm 0.000\degree$/s), and the rollout does not issue a stop signal, running to the length cap.

Switching to probabilistic mixture heads alone does not secure reliable termination at any tested temperature. The Prob.\ metrics in Table~\ref{tab:results_site_a} are teacher-forced and therefore temperature-independent. Under a common 500-sample horizon, above Site~A's 99th-percentile trip length of 491~s but below its 1,737~s maximum, the canonical IKTM pool stops naturally on 98 of 100 rollouts. The corresponding count across the duration-free sweep $T \in \{0.01, 0.1, 0.3, 0.5, 0.7, 0.9\}$ is 0--50 of 100. Both figures are censored from pools generated under the same 1,250-sample cap, so the two arms are compared at a common horizon over commonly generated pools.
Extending every duration-free arm to IKTM's 1,250-sample cap yields 0\% natural stops at $T \le 0.5$, 14\% at $T=0.7$, and 78\% at $T=0.9$; IKTM reaches 100\% at every tested temperature. Even at $T=0.9$, 22 of 100 duration-free rollouts hit the cap, and the natural-stop length p95 is 1,062 samples. Higher-temperature sampling increases eventual stopping, but termination remains temperature-sensitive, late, and incomplete rather than target-controlled. The individual $T=0.5$ sample likewise reaches sample 500 without a stop signal (terminal speed 13.0~m/s; terminal stop probability 0.000).
At low temperatures the duration-free probabilistic variant reverts to fully fixed-point-like behaviour over the 100-rollout diagnostic: at $T=0.01$ every rollout hits the 1,250-sample cap without a natural stop, at near-constant kinematics (peak speed ${\approx}12.9$~m/s, mean $|\Delta\theta| \approx 0.13\degree$/s). Across the whole range tested, temperature is not a substitute for duration conditioning (Section~\ref{sec:duration}).
The opposite ablation completes the $2\times2$ grid in the generative direction and separates the two mechanisms' contributions.
The deterministic duration-conditioned variant (Det.+Dur.), evaluated under exactly the protocol used for the full IKTM below ($N=100$ rollouts from BOS, Site~A test split, seed 42, same 1,250-step cap), also terminates on 100 of 100 rollouts with a length-tracking error of $+0.0 \pm 0.0$~s.
Duration conditioning alone therefore provides exact termination under this protocol without probabilistic heads. Its turn-rate JSD is 0.237~bits against the Site~A test distribution, 6.2$\times$ the full IKTM's 100-seed oracle-arm mean of 0.0385 and close to the persistence and prior-Markov baselines of Section~\ref{sec:baselines} (0.245 and 0.231). The two mechanisms address different outcomes: duration conditioning controls termination, while the probabilistic mixture heads reduce turn-rate JSD when the deterministic variant already stops on cue.
For the full IKTM (probabilistic heads + duration conditioning), we swept the sampling temperature $T \in \{0.01, 0.02, \dots, 0.09, 0.1, 0.2, \dots, 0.9\}$ with $N=100$ rollouts each, using the same 1,250-step cap as the canonical rollout-quality evaluation above so that no rollout is cap-truncated. This sweep is run on the Site~A \emph{validation} split, never the held-out test split, per the same split discipline used for checkpoint selection.
All temperatures achieve a 100\% natural stop rate (Wilson 95\% confidence interval on $100/100$: 96.3--100\%; the same bound applies to every other $N{=}100$, 100\% natural-stop figure reported in this paper). Against the Site~A validation distribution, turn-rate JSD is U-shaped in $T$, with a minimum of 0.041~bits at $T=0.2$. Near-greedy decoding under-disperses the heading mixture (0.25~bits at $T=0.01$), while higher temperatures over-disperse it (0.07--0.08~bits at $T \ge 0.6$). For this checkpoint and seed, both extremes increase JSD relative to $T=0.2$.
Length-tracking error std stays in a tight band across the whole sweep ($\le 0.22$~s at every temperature), so it does not meaningfully discriminate between temperatures. This reflects a small minority of $\pm1$~s off-target rollouts rather than a uniformly tight per-rollout error: the exact-on-target fraction is 100\% at every swept temperature except 99\% at $T=0.07$ and $T=0.08$, 98\% at $T=0.6$--$0.8$, and 95\% at $T=0.9$.
We therefore select $T=0.2$ as the default because it minimises turn-rate JSD against the Site~A validation reference (0.041~bits). Sampled speeds are additionally bounded by the data-derived 13.9~m/s ceiling (Section~\ref{sec:prob_heads}). This selection uses only the in-distribution Site~A validation split; Sites~B--D and the held-out Site~A test split are not consulted.

With the full IKTM at $T=0.2$, termination is reliable and length tracking is exact on the held-out test split: all 100 rollouts terminate naturally via the stop head, with a length error of $+0.0 \pm 0.0$~s against the sampled target (100 of 100 rollouts exactly on target).
Most sampled targets fall below the remaining-time feature's 300~s saturation ceiling (Section~\ref{sec:duration}), so a separate 600-rollout pool (seeds 41--46, same split and temperature) is additionally stratified by target duration.
Tracking is exact in both regimes: 565 of 565 rollouts with $D \le 300$~s and 35 of 35 with $D > 300$~s terminate exactly on target, the latter spending a median 19\% (up to 60\%) of their length with $\tau_t$ at the ceiling, on targets reaching 751~s.
The ceiling therefore limits what the countdown provides early in a long rollout, but not termination accuracy once the countdown takes effect. No saturation effect was measurable in the group of 35 rollouts.
The turn-rate JSD of 0.039~bits against the Site~A test distribution (Table~\ref{tab:rollout}) quantifies the residual gap between generated rollouts and the empirical marginal over the Site~A test devices' operating conditions. Generated rollouts always start from rest, approach rest by the stop event (the stop head does not impose an exact zero-speed constraint), and follow the learned duration prior.

\begin{figure}[!t]
\centering
\includegraphics[width=\columnwidth]{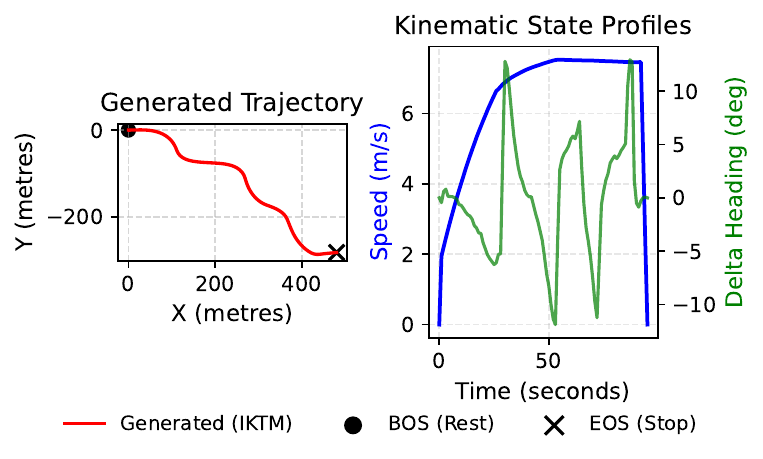}
\caption{An illustrative rollout from the full IKTM at $T=0.2$ (left) alongside its speed and heading-change profile (right). The black circle marks the BOS start, the black $\times$ the end-of-sequence (EOS) stop event.}
\label{fig:rollout_sample}
\end{figure}

Fig.~\ref{fig:rollout_sample} shows one such rollout: a complete, self-terminating trajectory with a smoothly varying speed and heading-change profile, in contrast to the non-terminating, speed-pinned rollouts produced by the deterministic baseline described above.
Its left panel is an integrated relative-frame path, obtained by cumulatively integrating the generated speed and heading-change sequence; it is not a direct model output, since IKTM predicts no coordinates.
The draw terminates via the stop head after 95~s, matching its own sampled target duration of 95~s exactly. Its length sits above the Site~A \emph{empirical} median trip length of 74~s (distinct from the \emph{fitted} log-duration prior's median of 80~s, Section~\ref{sec:duration}) and within the 95th percentile of 311~s.
This illustrative draw uses test-time symmetrised sampling: at each step the model is run on the context and its mirror, with the heading sign negated, and a fair coin drawn independently per step and per rollout picks the branch.
This removes the small residual left/right net-rotation skew of plain autoregressive sampling, at ${\sim}2\times$ forward cost.
The quantitative diagnostics in this section, meaning the $N=100$ pool and Table~\ref{tab:rollout}, were generated with plain sampling throughout, as was every other pool reported in this paper. Symmetrisation is therefore confined to this figure and enters no reported number.

\begin{table}[!t]
\caption{Generative Rollout Quality: IKTM ($N=100$ Rollouts per Site, Canonical Seed 42)}
\label{tab:rollout}
\centering
\scriptsize
\begin{threeparttable}
\begin{tabular}{llcccc}
\toprule
 & & \multicolumn{2}{c}{Oracle-length arm} & \multicolumn{2}{c}{Prior-length arm} \\
\cmidrule(lr){3-4} \cmidrule(lr){5-6}
Site & Type & Trip-len & Turn-rate & Trip-len & Turn-rate \\
     &      & JSD (bits) & JSD (bits) & JSD (bits) & JSD (bits) \\
\midrule
Site~A & ID-test & 0.061 & 0.039 & 0.061 & 0.039 \\
Site~B & OOD    & 0.047 & 0.057 & 0.081 & 0.044 \\
Site~C & OOD    & 0.049 & 0.031 & 0.060 & 0.027 \\
Site~D & OOD    & 0.024 & 0.033 & 0.039 & 0.036 \\
\bottomrule
\end{tabular}
\begin{tablenotes}
\small
\item Oracle-length arm: Site~A duration targets are drawn from the learned prior; Sites~B--D targets are drawn from each site's own empirical trip-length distribution (a 50-bin histogram over $[0, 1200]$~s, with the target set to the sampled bin's centre rather than an individual observed trip length), isolating kinematic quality from the Site~A-fitted prior. Prior-length arm: all sites, including B--D, draw duration targets from the Site~A-fitted prior alone (fully zero-shot in the duration dimension); Site~A's columns are therefore identical by construction. Both arms bin trip length on the same 50-bin grid over $[0, 1200]$~s.
\item JSD: 0 = identical distributions, 1 = maximally different.
\end{tablenotes}
\end{threeparttable}
\end{table}

Table~\ref{tab:rollout} extends the generative evaluation to the OOD sites, under two duration-target protocols, at the canonical illustrative seed (42); the 100-seed means reported below are used as the headline values.
Under the oracle-length arm, turn-rate JSD clusters at ${\approx}$0.031--0.057~bits across all four sites at the canonical seed, bracketing the in-distribution reference of 0.039~bits.
Trip-length JSD is low everywhere (0.024--0.061~bits). For Sites~B--D it is fixed by construction, given the exact length tracking reported above: targets are drawn as centres of the same 50-bin grid the JSD is then scored on, so every generated length re-enters the bin it was drawn from and the cell measures the $N{=}100$ multinomial resampling noise of that draw, carrying no model-dependent information. It confirms only that rollouts execute the requested durations as intended. For Site~A, whose targets come from the learned prior rather than from the site's own histogram, it shows that the learned log-duration prior itself reproduces the held-out test-split length distribution.

Under the fully zero-shot \emph{prior-length} arm, all four sites draw duration targets from the Site~A-fitted prior alone, so no held-out-site length information reaches the generator. Turn-rate JSD stays in the same band (${\approx}$0.027--0.044~bits at the canonical seed), closely comparable to the oracle-length arm, while trip-length JSD on Sites~B--D is higher by $+0.012$ to $+0.030$~bits when the two arms are paired per seed, with per-site standard deviations of 0.016--0.022~bits on those same differences. This increase is of the order of its own seed-to-seed spread; Site~A's columns are identical across arms by construction.

JSD is the distributional-realism measure also adopted by TrajGDM~\cite{chu2023trajgdm}, though over different movement statistics and datasets there.

A single-seed JSD estimate is itself a sample statistic: each rollout pool has only $N{=}100$ trajectories against millions of real steps.
We therefore repeated the full OOD generative evaluation over 100 sampling seeds, holding $T=0.2$ fixed throughout.
The sweep advances all four sites' rollouts as one batch per forward pass rather than one trajectory at a time, which is what makes 100 seeds tractable. Every per-step operation matches the sequential generator used for Table~\ref{tab:rollout}, but the batch draws from the same distributions in a different order, so a given seed yields an independent draw rather than a replay of the sequential one; the sweep script also provides a sequential mode that reproduces the sequential output exactly.
The resulting mean$\pm$std is the headline turn-rate JSD figure quoted in the abstract and Conclusions, rather than the single canonical-seed snapshot in Table~\ref{tab:rollout}.
Across these 100 seeds, turn-rate JSD is $0.0385 \pm 0.0047$ (Site~A), $0.0502 \pm 0.0063$ (Site~B), $0.0357 \pm 0.0052$ (Site~C), and $0.0354 \pm 0.0037$ (Site~D) (mean $\pm$ sample std, $N=100$ rollouts per seed); trip-length JSD over the same seeds is $0.0565 \pm 0.0129$ (Site~A), $0.0623 \pm 0.0121$ (Site~B), $0.0597 \pm 0.0132$ (Site~C), and $0.0392 \pm 0.0088$ (Site~D).
The four per-site means span 0.035--0.050~bits, whose endpoints are both unseen sites: the widest, Site~B at 0.0502~bits, is 1.30$\times$ the in-distribution mean of 0.0385~bits.

Under the prior-length arm, across the same 100 seeds, turn-rate JSD is $0.0388 \pm 0.0051$ (Site~A), $0.0423 \pm 0.0063$ (Site~B), $0.0324 \pm 0.0040$ (Site~C), and $0.0385 \pm 0.0042$ (Site~D). These values overlap the oracle-length-arm range, so the measured turn-rate transfer does not depend on using each site's length data for the duration target. Trip-length JSD under the prior-length arm is $0.0565 \pm 0.0129$ (Site~A, identical to the oracle-length arm by construction), $0.0773 \pm 0.0150$ (Site~B), $0.0894 \pm 0.0173$ (Site~C), and $0.0511 \pm 0.0142$ (Site~D).
Turning fidelity is therefore stable to a per-site standard deviation of at most 0.0063~bits across seeds, and the choice of duration-target arm registers in trip length rather than in turn rate.

\subsection{External Baselines and the JSD Noise Floor}
\label{sec:baselines}

To situate the turn-rate JSD numbers above, we score three of Table~\ref{tab:baselines}'s reference points (rows 2--4) under a closely matched protocol: $N=100$ rollouts per site, seed 42, and oracle-length duration targets throughout, since none of these baselines has a learned duration mechanism of its own.
The three are as follows: our prior Markovian mobility model~\cite{Amiri2024Model}, re-implemented as a standalone rollout generator with four speed-class-conditioned exponential direction- and speed-change models with one-step memory (Section~\ref{sec:related_industrial}), using that paper's own published per-class parameters rather than refitting them here; an i.i.d.-marginal sampler that draws each per-step kinematic value independently from the real marginal, with no temporal structure whatsoever; and a persistence/zero-turn baseline ($\Delta\theta \equiv 0$ every step).

The baselines follow this protocol at all four sites.
The IKTM figures these are compared against are the 100-seed oracle-arm means reported in Section~\ref{sec:rollout_quality}, not a row of Table~\ref{tab:baselines}; at Site~A they draw duration targets from the learned prior rather than Site~A's empirical histogram, per that arm's convention. The two sources are closely matched: fitted prior median 79.9~s against empirical median 74~s (Section~\ref{sec:duration}).
Amiri et al.\ report only unsigned direction- and speed-change magnitudes, so the reimplementation assumes a symmetric coin flip for speed-change sign, clipped at zero since speed cannot go negative.
The reimplementation advances once per generated second. Its published per-class change magnitudes were estimated over the source telematics' irregular inter-fix intervals (median 5~s, 99th percentile 45~s; Section~\ref{sec:introduction}). Both generators are scored at the 1~Hz resolution used by IKTM and the metric. The comparison therefore measures per-second kinematic accuracy under this common protocol, not the prior model's fit at its original event spacing.
The sign affects future speed-class occupancy and therefore the class-conditioned unsigned turn distribution. In stream-controlled sensitivity arms that still consume the same sign coin at every step, forcing acceleration or deceleration moves the per-site turn-rate JSD range from the canonical 0.197--0.267~bits to 0.190--0.306~bits; IKTM remains 5.1--6.8$\times$ better across the two extreme policies when paired per site.
We additionally establish the metric's own sampling-noise floor by bootstrap-resampling $N=100$ \emph{whole real trips}, with replacement, preserving each trip's own length and turn sequence together, over 500 repetitions at seed 42.
Whole trips match the actual statistical unit the headline JSD pools, namely steps from 100 dependent, within-trip-autocorrelated trajectories; resampling 100 independent steps instead would understate the floor's true sampling variance.

\begin{table*}[t]
\caption{Turn-Rate JSD Baselines (Matched Protocol, $N=100$, Seed 42 Unless Noted)}
\label{tab:baselines}
\centering
\scriptsize
\begin{threeparttable}
\begin{tabular}{lcccc}
\toprule
 & Site~A & Site~B & Site~C & Site~D \\
\midrule
Noise floor (mean, 95\% bootstrap interval) & 0.0016 [0.0008,0.0034] & 0.0014 [0.0007,0.0033] & 0.0019 [0.0007,0.0049] & 0.0017 [0.0009,0.0034] \\
i.i.d.\ marginal            & 0.0006 & 0.0007 & 0.0009 & 0.0011 \\
Persistence (zero-turn)     & 0.245  & 0.211  & 0.287  & 0.275 \\
Prior Markov~\cite{Amiri2024Model} & 0.231 & 0.267 & 0.202 & 0.197 \\
IKTM trained on linear interp.\ (Section~\ref{sec:bezier}) & 0.233 & 0.198 & 0.274 & 0.256 \\
\textbf{IKTM (ours)}, 100-seed mean & \textbf{0.039} & \textbf{0.050} & \textbf{0.036} & \textbf{0.035} \\
\bottomrule
\end{tabular}
\begin{tablenotes}
\small
\item Cells are turn-rate JSD (bits); for these rows, trip-length JSD spans 0.02--0.08~bits. The three non-IKTM generators draw empirical duration targets at every site; the linear-trained IKTM and the IKTM row both follow Table~\ref{tab:rollout}'s oracle-length-arm convention, using the learned prior at Site~A and empirical targets at Sites~B--D. The IKTM row is the 100-seed mean of Section~\ref{sec:rollout_quality} (seeds 1--100); every other row is the single seed-42 draw of the caption. The corresponding whole-trip bootstrap values appear below; turn-rate JSD is the metric that discriminates between these baselines.
\end{tablenotes}
\end{threeparttable}
\end{table*}

The upper end of the noise floor's 95\% bootstrap interval (the 97.5th percentile of the 500 resampled JSD values) is at most 0.0049~bits across the four sites, below IKTM's 100-seed mean turn-rate JSD of 0.032--0.050~bits (Section~\ref{sec:rollout_quality}). The prior Markov model scores 0.197--0.267~bits under the canonical symmetric-sign policy, 5.3--6.0$\times$ IKTM's paired per-site oracle-arm means. The extreme sign policies above produce the same ordering. The persistence/zero-turn baseline scores 0.211--0.287~bits, showing that the metric separates this baseline from IKTM as well.
The same bootstrap gives trip-length JSD means of 0.041--0.063~bits per site, with site-specific 95\% bootstrap-interval upper bounds of 0.062--0.092~bits. The reported trip-length JSDs (0.024--0.089~bits across both duration-target arms and all four sites) are of the same order as this $N{=}100$ whole-trip resampling variability; turn-rate JSD is the column that separates generators here.

The i.i.d.-marginal baseline scores turn-rate JSD of 0.0006--0.0011~bits at the noise floor because drawing each step independently from the real marginal matches the pooled turn-rate distribution by construction. The persistence baseline's 0.211--0.287~bit score shows that the metric responds to a different marginal.

Turn-rate JSD certifies a matched marginal distribution, not temporal realism. Table~\ref{tab:temporal} therefore compares IKTM and the i.i.d.-marginal sampler on two temporal statistics already used in this paper. Both use duration targets drawn from each site's own trip-length histogram under the oracle-length mechanism above; only the within-rollout kinematic generator differs.
The sign-change rate between consecutive nonzero $\Delta\theta$ steps places the i.i.d.\ sampler at 0.494--0.507 across the four sites, consistent with the 0.5 expected of uncorrelated jitter under the null that Section~\ref{sec:bezier} adopts, against 0.176--0.201 for real data and 0.158--0.169 for IKTM.
Per-second acceleration separates the arms by three orders of magnitude: the i.i.d.\ sampler's p99 is 10.3--15.7~m/s\textsuperscript{2}, with 22.0--40.6\% of its steps above the 3~m/s\textsuperscript{2} bound used in Section~\ref{sec:robustness}, against 0.004--0.008\% of real steps and 0.000--0.044\% of IKTM's.
A matched marginal is thus compatible with per-step dynamics far from those of the real data, and these two statistics distinguish the arms that turn-rate JSD does not.
IKTM's sign-change rate lies slightly below the real value at every site (0.158--0.169 against 0.176--0.201), indicating turning marginally more persistent than in the real data, in the same direction as the total-turning comparison in Section~\ref{sec:robustness}.

\begin{table}[!t]
\caption{Temporal Structure Under a Matched Protocol ($N=100$, Seed 42)}
\label{tab:temporal}
\centering
\scriptsize
\begin{threeparttable}
\begin{tabular}{lcccc}
\toprule
 & Site~A & Site~B & Site~C & Site~D \\
\midrule
\multicolumn{5}{l}{\emph{Sign-change rate between consecutive nonzero $\Delta\theta$ steps}} \\
Real                        & 0.201 & 0.176 & 0.194 & 0.201 \\
i.i.d.\ marginal            & 0.502 & 0.494 & 0.504 & 0.507 \\
\textbf{IKTM}               & 0.162 & 0.158 & 0.169 & 0.163 \\
\midrule
\multicolumn{5}{l}{\emph{Per-second acceleration exceeding 3~m/s\textsuperscript{2} (\% of steps)}} \\
Real                        & 0.008 & 0.004 & 0.004 & 0.008 \\
i.i.d.\ marginal            & 39.87 & 21.98 & 40.58 & 34.33 \\
\textbf{IKTM}               & 0.044 & 0.009 & 0.000 & 0.012 \\
\bottomrule
\end{tabular}
\begin{tablenotes}
\small
\item 0.5 is the sign-change rate expected of uncorrelated jitter (Section~\ref{sec:bezier}); the 3~m/s\textsuperscript{2} bound is the one cross-checked against real data in Section~\ref{sec:robustness}. The real rows are per-site over the same device filter as the generated arms (Site~A: test split), so they are not the four-site pooled figures quoted in Section~\ref{sec:robustness}. This pool is drawn fresh under the matched protocol and is not a bit-for-bit replay of the headline evaluation's rollouts.
\end{tablenotes}
\end{threeparttable}
\end{table}

Turn-rate JSD therefore separates these generators into three bands: the i.i.d.\ sampler at 0.0006--0.0011~bits, within the metric's ${\le}0.0049$~bit floor; IKTM at 0.035--0.050~bits; and the Markov, persistence, and linear-trained baselines at 0.197--0.287~bits.

\subsection{Robustness of the Distributional Match}
\label{sec:robustness}

Five analyses examine the headline JSD through distributional distance, bin count, speed conditioning, signed turning, and the speed marginal, followed by a physical-plausibility check. They use a fresh $N=100$ pool per site drawn under Table~\ref{tab:rollout}'s duration-target protocol with the same checkpoint, $T=0.2$, and seed 42. The Site~A pool is bit-identical to Table~\ref{tab:rollout}'s canonical-seed evaluation because both scripts process it first after resetting the random state. The Sites~B--D pools are independently sampled: the headline evaluation uses one continuous random stream across sites, whereas this diagnostic resets the seed per site and draws duration targets from an independent generator. Turn-rate JSD is 0.024--0.044~bits in these pools, compared with 100-seed oracle-arm means of 0.035--0.050~bits. Site~C's new pool scores 0.024~bits, at the low end of its 100-seed range of 0.023--0.051. The following analyses therefore characterise these pools rather than replicate Table~\ref{tab:rollout}'s Sites~B--D cells.

\textbf{Beyond 50-bin JSD.} A two-sample, two-sided Kolmogorov--Smirnov (KS) test (SciPy's \texttt{ks\_2samp}, default asymptotic method) between generated and real $|\Delta\theta|$ yields KS $=0.36$--$0.47$ and $p \ll 0.001$ at all four sites. The corresponding Wasserstein distance is 4.2--7.1\degree. These measures expose differences not expressed by the 50-bin JSD values of 0.024--0.044~bits.
The test pools steps from within-trip-autocorrelated trajectories rather than independent draws, so the reported $p$-value is indicative rather than an exact tail probability. The KS statistic of 0.36--0.47 and the Wasserstein distance quantify the measured separation independently of that nominal tail probability.
Re-binning the same samples at 10/25/50/100/200 bins increases JSD monotonically from 0.012--0.020 (10 bins) to 0.11--0.16 (200 bins). Thus, the numerical value depends on bin count, and finer bins expose more small-scale mismatch.
At 50 bins, IKTM scores 0.024--0.044~bits on these pools, below the linear, Markov, and persistence baselines' 0.197--0.287~bit ranges but above the i.i.d.\ marginal sampler's 0.0006--0.0011~bits. The temporal statistics in Table~\ref{tab:temporal} distinguish IKTM from that sampler.

\textbf{Speed-conditional turning.} Splitting $|\Delta\theta|$ by the same four speed classes as the Markov baseline (Section~\ref{sec:related_industrial}), turn-rate JSD is 0.022--0.053~bits at all four sites in the two classes with adequate sample support (1--19 and 20--39~km/h), comparable to the unconditional headline.
The higher classes are not reported. IKTM's 13.9~m/s generation ceiling is 50~km/h, so the 60+~km/h class is unreachable by construction and 40--59~km/h only partly reachable; that, together with the low-speed operating profile, leaves too little rollout mass above 40~km/h for a stable estimate at this $N$.

\textbf{Speed marginal.} We evaluate the second generated channel on the same pools, binning per-step speed into 50 bins over $[0, 25]$~m/s. This range includes speeds above the 13.9~m/s generation ceiling, so the comparison retains real observations that the model cannot generate.
The residual mass beyond the 25~m/s upper edge is excluded on both sides, but that is 0.000\% of real steps at Site~A and at most 0.020\% at Sites~B--D, two orders of magnitude below the turn-rate histogram's excluded tail. Speed-marginal JSD is 0.071 (Site~A), 0.129 (Site~B), 0.081 (Site~C) and 0.098 (Site~D)~bits, 1.8--3.4 times the turn-rate JSD, and larger at every site.
Two mechanisms account for the difference. First, the generated median speed exceeds the real median at all four sites (5.39 against 4.46~m/s at Site~A): a rollout is a single trip from rest to rest, whereas the real per-step pool also contains extended low-speed intervals within trips. Second, on the held-out sites the generated speed scale follows Site~A's rather than the host site's: generated p95 is 9.12~m/s at Site~B against a real 6.94~m/s, and 8.76~m/s at Site~C against a real 11.72~m/s.
The 13.9~m/s clamp fitted on Site~A's train and validation devices (Section~\ref{sec:prob_heads}) compounds this, placing 3.1\% of Site~C's real steps outside the reachable range, against 0.35\% of Site~A's. Site~C also has the largest fast-tier share of the four fleets (16.5\% against Site~A's 1.8\%, Table~\ref{tab:dataset_stats}), so the difference corresponds to fleet composition as well as to the clamp.
The two evaluations measure different quantities. Next-step speed RMSE is conditional on the true recent history and remains 0.13--0.15~m/s OOD. The free-running speed marginal is unconditional and follows Site~A's operating profile because the model inputs include a target duration but no target speed scale. The transfer result therefore applies to conditional manoeuvre prediction, not to the host site's unconditional speed regime.

\textbf{Signed turning and net rotation.} The real per-step signed $\Delta\theta$ mean is small at every site (0.04--0.22\degree), as expected for an unbiased population.
Generated rollouts show a signed mean of $-0.01$ to $0.55\degree$/step, largest at Site~C and larger than the real value at three of the four sites, which accumulates into a non-trivial mean net rotation per rollout over the rollout's full length: Site~A $+30\pm126$\degree, Site~B $-1\pm132$\degree, Site~C $+56\pm133$\degree, and Site~D $+18\pm102$\degree.
This is a small but real left/right imbalance in plain autoregressive sampling, and is already the motivation for the test-time symmetrisation used only for the qualitative figure (Fig.~\ref{fig:rollout_sample}).
The per-site standard deviations of 102--133\degree{} exceed the corresponding mean rotations of $-1$ to $+56\degree$, although the mean effect is nonzero.

\textbf{Physical plausibility.} Across all 400 generated rollouts (4 sites $\times$ $N{=}100$), per-second acceleration has $|a|$ p50 $=0.006$ and p99 $=1.94$~m/s\textsuperscript{2}; 0.005\% of steps exceed 3~m/s\textsuperscript{2}.
Jerk has p99 $=1.52$~m/s\textsuperscript{3}, with 0\% of steps above 3~m/s\textsuperscript{3}. Curvature has p99 $=0.243$~rad/m, corresponding to an approximately 4~m turning radius, with 0.503\% of steps above 0.5~rad/m. Curvature is defined as $|\Delta\theta|$ per metre travelled and is excluded below 0.5~m/s. The three bounds are fixed a priori as round physical limits for a low-speed service vehicle, then cross-checked against real per-trip data under identical site and device filters ($N=81{,}795$ trips).
Real data exceeds each bound at a higher per-step rate than generated rollouts: 0.006\% against 0.005\% for acceleration, 0.015\% against 0.000\% for jerk, and 1.609\% against 0.503\% for curvature. Terminal speed at the stop event has a mean of 0.002~m/s and p95 of 0.0~m/s.
Together these lenses scope the headline claim: a small-turn agreement, measured at a fixed bin count, that also holds when split by speed class, over rollouts staying inside the acceleration, jerk, and curvature envelope of real trips.

The final row of Table~\ref{tab:baselines} is a controlled end-to-end ablation of the curved-interpolation contribution (Section~\ref{sec:bezier}), using an otherwise-identical IKTM trained on linearly interpolated Site~A timeseries. Section~\ref{sec:bezier} compares the two interpolation schemes on input statistics, namely the zero-$\Delta\theta$ fraction and per-trip total turning; this ablation measures their effect on generated output.
The linear-trained model's turn-rate JSD is 0.198--0.274~bits, comparable to the prior Markov baseline and, paired per site against the curved-trained IKTM's own 100-seed oracle-arm means, 3.9--7.7$\times$ higher.
Because the JSD reference is the curved export, we also compare median total absolute turning with the raw checkpoint record, which contains no interpolated headings. Curved training raises the generated median from 80\degree{} to 360\degree{}, crossing the raw-record median of 260\degree{}; the curved-trained value is 1.4$\times$ the raw record and 46\% of the curved export's 786.0\degree{} median (Section~\ref{sec:bezier}). The generated rollouts have a realised median length of 89~s against the real trips' 74~s empirical median, so the populations are comparable but not duration-matched. Accordingly, the ablation establishes the direction of the interpolation effect on generated turning, with its 360\degree{} median remaining below the curved export's 786.0\degree{}.
This trip-level figure corresponds to the per-step distribution reported in Section~\ref{sec:metrics}, where generated rollouts place 0.9--1.0\% of their steps beyond $45\degree$ against 4.5--6.8\% of real steps: the smaller large-turn mass per step accumulates into a smaller total per trip.

\section{Conclusions and Future Work}
\label{sec:conclusions}

We presented the Industrial Kinematic Trajectory Model (IKTM), a coordinate-free autoregressive trajectory generator for unstructured industrial environments.
IKTM addresses three design requirements: (i) preserving turns in sparse GPS data, via heading-aware cubic B\'{e}zier interpolation; (ii) varying open-loop trajectories, via probabilistic GMM and von Mises mixture heads; and (iii) controlling rollout length, via duration conditioning with a learned log-duration prior.
IKTM is trained on a single site and evaluated on real telematics from four.
Across all four sites, with Sites~B--D held out entirely zero-shot, the rollouts generate turn-rate distributions whose per-site mean JSD over 100 sampling seeds spans ${\approx}$0.035--0.050~bits under an oracle-length duration-target protocol and ${\approx}$0.032--0.042~bits under a fully zero-shot prior-length protocol.
That is lower than a re-implementation of our own prior Markovian model, scored at 1~Hz under the same protocol (turn-rate JSD 0.197--0.267~bits under the canonical sign policy, Section~\ref{sec:baselines}), while remaining above the metric's sampling-noise floor ($\le$0.0049~bits): small-turn transfer with a measurable remaining difference and no geographic coordinate exposure.
Termination is duration-conditioned rather than spatial: rollouts stop on 100\% of trials with a length-tracking error of $+0.0 \pm 0.0$~s, measured in the in-distribution rollout diagnostic ($N=100$, $T=0.2$, held-out test split), with the target duration sampled from the model's own learned prior rather than leaked from ground truth.
This extends our previous Markovian industrial mobility model~\cite{Amiri2024Model} to a deep autoregressive setting, replacing single-step speed-class transitions with a 60-step continuous kinematic context and adding explicit trip-termination control.

The results suggest that industrial trajectory models require a distinct design point from universal trajectory foundation models~\cite{zhu2025unitraj,han2025movegpt}: coordinate-free kinematics, industrial-specific preprocessing, and explicit termination control.

\textbf{Scope.}
IKTM is trained on one site and evaluated on four. Under both duration-target arms, the 100-seed per-site mean turn-rate JSD spans 0.032--0.050~bits across the in-distribution and zero-shot sites, while leave-one-site-out next-step evaluation finds no systematic home-versus-cross-site gap. The open-loop speed marginal instead follows Site~A's operating profile. Four sites do not establish that the same manoeuvre structure holds for every industry and vehicle class. The model is one instantiation of coordinate-free kinematic autoregression and exposes no absolute coordinates.

\textbf{Future work} includes:
(i) conditioning generation on semantic zone maps of industrial sites, enabling origin--destination-constrained simulation similar to MoveFM-R's counterfactual reasoning~\cite{meng2026movefmr};
(ii) multi-class extension for heterogeneous vehicle fleets (forklifts, haul trucks, reach stackers) using a Mixture-of-Experts architecture inspired by MoveGPT~\cite{han2025movegpt};
(iii) self-supervised masked kinematic pre-training as a complement to the current causal objective, following insights from UniTraj (Zhu et al.)~\cite{zhu2025unitraj} and TrajTok~\cite{xiong2026trajtok}.

\textbf{Reproducibility and source availability.} The canonical trained IKTM checkpoint is publicly available under the MIT License at \url{https://huggingface.co/maxamiri/iktm}. The full preprocessing, training, and evaluation pipeline, together with its associated scripts, may be released in the future. The underlying telematics data cannot be released because the industrial site agreements designate it as confidential. The same agreements require the site anonymisation used throughout the paper (Section~\ref{sec:evaluation}).

\bibliographystyle{IEEEtran}
\bibliography{ref}

\end{document}